\documentclass{article} 
\usepackage{iclr2025,times}

\usepackage{amsmath,amsfonts,bm}

\def\eqref#1{equation~\ref{#1}}

\def\1{\bm{1}}

\DeclareMathAlphabet{\mathsfit}{\encodingdefault}{\sfdefault}{m}{sl}
\SetMathAlphabet{\mathsfit}{bold}{\encodingdefault}{\sfdefault}{bx}{n}

\usepackage{hyperref}
\usepackage{url}
\usepackage{graphicx}
\usepackage{subfigure}
\usepackage{booktabs}

\usepackage{amsmath}
\usepackage{amssymb}
\usepackage{mathtools}
\usepackage{amsthm}

\usepackage{multirow}
\usepackage{color}
\usepackage{colortbl}
\usepackage[capitalize,noabbrev]{cleveref}
\usepackage{xspace}
\usepackage{etex}
\usepackage[etex=true,export]{adjustbox}
\usepackage{url}
\usepackage{marvosym}

\definecolor{emphypurple}{rgb}{0.302, 0.055, 0.659}
\usepackage{hyperref}
\usepackage{url}
\usepackage{hyperref}
\usepackage{xcolor}
\usepackage{pifont}
\usepackage{soul}
\usepackage{url}
\usepackage[T1]{fontenc}
\usepackage{siunitx}
\usepackage{graphicx}
\usepackage{float}
\usepackage{multirow}
\usepackage{color}
\usepackage{wrapfig}
\usepackage{booktabs}
\usepackage{amssymb}
\usepackage{subfigure}
\usepackage{listings}
\usepackage{makecell}
\usepackage{amssymb,mathrsfs,amsmath}
\usepackage{amsmath}
\usepackage{colortbl}
\usepackage{xcolor}
\usepackage{lipsum}
\usepackage{comment}

\usepackage{enumitem}

\usepackage[utf8]{inputenc} 
\usepackage[T1]{fontenc}    
\usepackage{hyperref}       
\usepackage{url}            
\usepackage{booktabs}       
\usepackage{amsfonts}       
\usepackage{nicefrac}       
\usepackage{microtype}      
\usepackage{xcolor}         
\usepackage{csquotes}
\definecolor{warningcolor}{RGB}{255, 0, 0}

\definecolor{xred}{HTML}{BD4242}
\definecolor{xblue}{HTML}{C7A085}
\definecolor{xblues}{HTML}{52B256}
\definecolor{xgreen}{HTML}{52B256}
\definecolor{xpurple}{HTML}{7F52B2}
\definecolor{xorange}{HTML}{FD9337}
\definecolor{xdotted}{HTML}{999999}
\definecolor{xgray}{HTML}{777777}
\definecolor{xcyan}{HTML}{80F5DC}
\definecolor{xpink}{HTML}{f690ea}
\definecolor{xgraycyan}{HTML}{82bceb}
\usepackage{tikz}
\usetikzlibrary{positioning, calc}

\usepackage[most]{tcolorbox}

\usepackage{todonotes}
\usepackage{xcolor}

\tcbset{
	common/.style n args={2}{
		colframe={#1},
		colback={#1!5},
		colbacktitle={#1},
		lower separated=false,
		coltitle=white,
		boxrule=1pt,
		fonttitle=\bfseries,
		enhanced,
		breakable,
		top=8pt,
		before skip=8pt,
		attach boxed title to top left={
			yshift=-0.25cm,
			xshift=0.38cm,
		},
		boxed title style={
			boxrule=0pt,
			colframe=white,
			arc=5pt,
			outer arc=4pt,
		},
		separator sign={~~},
		overlay unbroken and last={
			\node[text=white, align=right, rounded corners, fill=#1, xshift=-4mm, minimum height=6mm, anchor=east] at (frame.south east) {#2};
		}
	},
	defstyle/.style={
		common={xpurple}{$\bm{\pi}$},
	},
	theoremstyle/.style={
		common={xgraycyan}{$\star$},
	},
	proofstyle/.style={
		common={xgreen}{\textbf{QED}},
	},
	examplestyle/.style={
		common={xblue}{\textbf{?}},
	},
    examplestyles/.style={
		common={xblues}{$\bigstar$},
	},
	notestyle/.style={
		common={xred}{$\bigstar$},
	},
	challangestyle/.style={
		common={xorange}{\textbf{?}},
	},
}

\theoremstyle{plain}

\theoremstyle{definition}

\theoremstyle{remark}

\definecolor{warningcolor}{RGB}{255, 0, 0}

\graphicspath{{./figures/}} 

\title{
T-Detect: Tail-Aware Statistical Normalization for Robust Detection of Adversarial Machine-Generated Text
}

\author{DeepScientist, \\
\textbf{Luodan Zhang}, \textbf{Liuliu Zhang}, \textbf{Minjun Zhu}, \textbf{Yixuan Weng}, \textbf{Yue Zhang} \\
School of Engineering, Westlake University\\
zhangyue@westlake.edu.cn}

\iclrfinalcopy
\begin{document}
\maketitle

\begin{abstract}
Large language models (LLMs) have shown the capability to generate fluent and logical content, presenting significant challenges to machine-generated text detection, particularly text polished by adversarial perturbations such as paraphrasing. 
Current zero-shot detectors often employ Gaussian distributions as statistical measure for computing detection thresholds, which falters when confronted with the heavy-tailed statistical artifacts characteristic of adversarial or non-native English texts.  
In this paper, we introduce T-Detect, a novel detection method that fundamentally redesigns the curvature-based detectors. Our primary innovation is the replacement of standard Gaussian normalization with a heavy-tailed discrepancy score derived from the Student's t-distribution. 
This approach is theoretically grounded in the empirical observation that adversarial texts exhibit significant leptokurtosis, rendering traditional statistical assumptions inadequate. T-Detect computes a detection score by normalizing the log-likelihood of a passage against the expected moments of a t-distribution, providing superior resilience to statistical outliers. 
We validate our approach on the challenging RAID benchmark for adversarial text and the comprehensive HART dataset. Experiments show that T-Detect provides a consistent performance uplift over strong baselines, improving AUROC by up to 3.9\% in targeted domains. When integrated into a two-dimensional detection framework (CT), our method achieves state-of-the-art performance, with an AUROC of 0.926 on the Books domain of RAID. Our contributions are a new, theoretically-justified statistical foundation for text detection, an ablation-validated method that demonstrates superior robustness, and a comprehensive analysis of its performance under adversarial conditions.  Ours code are released at \url{https://github.com/ResearAI/t-detect}.

{\color{warningcolor} \normalsize WARMING:
We hereby declare that ours DeepScientist system performed approximately 95\% of the work presented in this paper. This includes the initial ideation, the design and execution of comparative experiments, the analysis of results, the literature review, the composition of the manuscript, the creation of the main figures, and the organization of the accompanying code repository. The role of the human authors was to supervise the AI’s operations. While we have diligently worked to minimize AI hallucinations and ensure the validity of the experimental results, we cannot fully guarantee against potential unintended outputs, system failures, or misleading conclusions. We therefore advise readers to approach this work with caution and to critically evaluate its findings before application.}
 
\end{abstract}

\section{Introduction}

The rise of powerful large language models (LLMs) \citep{ouyang2022training,yang2025qwen3} has ignited a critical arms race between text generation and detection \citep{You2023LargeLM,Moraffah2024AdversarialTP}. While these models fuel innovation, they also carry risks like misinformation and academic dishonesty, making reliable detection essential \citep{Kumarage2024ASO}. However, this is not a static battlefield. A more dangerous front has opened: malicious actors are no longer just using LLMs, but are actively studying our detectors to craft adversarial attacks that can evade them \citep{You2023LargeLM, Lee2023PrompterSA}. These evolving strategies, from simple paraphrasing to subtle manipulations \citep{Li2024EnhancingTR}, demand a new generation of detectors built not just for accuracy, but for fundamental resilience.

The vulnerability of many current zero-shot detectors lies not on the surface, but deep in their statistical core. Leading methods like DetectGPT \citep{Mitchell2023DetectGPTZM} and Fast-DetectGPT \citep{Bao2023FastDetectGPTEZ} are built on a seemingly innocuous assumption: that their statistical scores follow a standard bell curve, or Gaussian distribution \citep{Rousseeuw2011RobustSF}. This is their Achilles' heel. Our empirical analysis reveals that adversarial texts are designed to break this premise. They produce score distributions with extreme outliers, resulting in "heavy-tailed" statistical properties \citep{dugan2024raid}. \textbf{The critical research problem, therefore, is that this violation of the Gaussian assumption makes detectors catastrophically sensitive to adversarial attacks, causing their performance to become unstable and unreliable.} When faced with the very texts they are designed to catch, their statistical foundation crumbles.

To this end, \textbf{we introduce T-Detect, a novel method that redesigns the detector's statistical core by replacing the flawed Gaussian assumption with a robust, "tail-aware" normalization based on the Student's t-distribution.} This single, principled change is grounded in robust statistics \citep{Rousseeuw2005RobustRA} and allows our method to gracefully handle the statistical outliers common in adversarial text without being destabilized. By computing a "heavy-tailed discrepancy score," T-Detect provides an inherently more stable and reliable signal for distinguishing human from machine-generated text.

\begin{figure*}[ht]
\centering
\includegraphics[width=\textwidth]{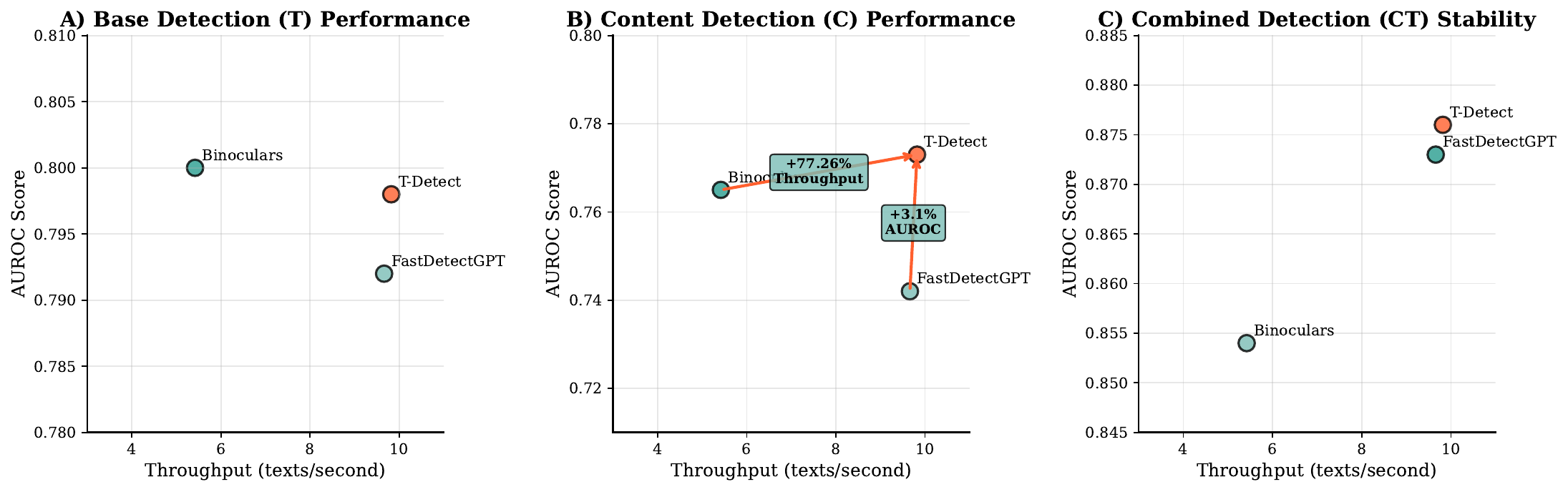}
\caption{The 'ALL' Performance (AUROC) vs. Speed (Throughput) on the RAID benchmark. T-Detect consistently provides a better Pareto frontier, offering higher performance for its computational cost. In the two-dimensional setting (c), CT(T-Detect) achieves state-of-the-art accuracy while being 1.8x faster than the competitive CT(Binoculars) baseline.}
\label{fig:perf_vs_speed}
\end{figure*}

We validate T-Detect through a comprehensive suite of experiments, demonstrating its practical advantages. As summarized in Figure~\ref{fig:perf_vs_speed}, T-Detect offers a superior trade-off between performance and computational efficiency compared to strong baselines. On the challenging RAID benchmark for adversarial text, our method, particularly when integrated into a two-dimensional (CT) framework\citep{bao2025decouplingcontentexpressiontwodimensional}, achieves state-of-the-art performance with an overall AUROC of 0.876. Our contributions are threefold: (1) We are the first to empirically prove that adversarial text detection scores follow heavy-tailed distributions and propose a theoretically-justified t-distribution-based normalization to address this. (2) We present an ablation-validated method that demonstrates superior robustness and performance on adversarial benchmarks. (3) We provide a comprehensive analysis of our method's practical benefits, including its computational stability and exceptional hyperparameter robustness, offering a more reliable and deployable solution for AI safety.

\section{Related Work}

The task of distinguishing machine-generated text from human-written content has evolved significantly, moving from early statistical methods to sophisticated zero-shot classifiers. Early approaches focused on identifying statistical artifacts in generated text. For instance, methods based on simple metrics like likelihood, log-rank, and entropy \citep{guo2023close,li2022artificial} were proposed to capture the unusually predictable nature of text from older generative models \citep{Gehrmann2019GLTRSD}. A significant breakthrough came with the introduction of curvature-based detection by \citet{Mitchell2023DetectGPTZM} in their seminal work, DetectGPT. This method was the first to hypothesize that text sampled from a large language model tends to occupy regions of high negative curvature in the model's log-probability space. DetectGPT estimated this curvature by generating numerous perturbations of a given text and measuring the average drop in log-probability, establishing a new paradigm for zero-shot detection that did not require a dedicated training dataset.

Building on this foundation, subsequent research has focused on improving both the efficiency and accuracy of curvature-based methods. Our direct baseline, Fast-DetectGPT, was introduced by \citet{Bao2023FastDetectGPTEZ} as a computationally efficient alternative to DetectGPT. It retains the core curvature hypothesis but replaces the costly perturbation step with a more efficient sampling-based approach to approximate the necessary statistics, achieving a significant speedup. Parallel to these developments, other zero-shot methods have emerged. Binoculars \citep{Hans2024SpottingLW} proposed a novel approach based on the cross-perplexity between two different language models, one acting as an "observer" and the other as a "performer." Another prominent method, NPR from the DetectLLM framework \citep{Su2023DetectLLMLL}, leverages log rank information, offering a different statistical signal for detection. Our work, T-Detect, contributes to the curvature-based lineage, but instead of focusing on computational efficiency, we address a more fundamental statistical limitation in the normalization step of these detectors.

To further enhance detection capabilities, some methods combine signals from multiple text representations, a common practice in the broader field of text classification \citep{Yang2013CombiningLA, Agarwal2014FrameST}. The two-dimensional (CT) detection framework, utilized in prior work, is one such approach. It combines a score from the original text (T) with a score from a content-only representation (C), where function words and other stylistic markers have been removed. This allows the system to decouple signals related to the expression of the text from those related to its core content. In our work, we use this framework to demonstrate that T-Detect provides a more robust base signal, thereby improving the performance of the entire combined system. This is particularly important in the context of adversarial attacks, such as paraphrasing \citep{Li2024EnhancingTR} and Unicode manipulation, which are designed to evade detection by altering either the expression or the underlying character data of a text, underscoring the need for robust, multi-faceted detection strategies.

\section{Method}
\label{sec:method}

The challenge of detecting machine-generated text has intensified with the advent of models capable of producing highly fluent and contextually appropriate content. A significant frontier in this field is the detection of text that has been adversarially perturbed to evade detection. Many existing zero-shot statistical detectors, such as Fast-DetectGPT \citep{Bao2023FastDetectGPTEZ}, operate by measuring the 'surprise' of a given text under a language model. They typically compute a discrepancy score representing how much the log-probability of the observed text deviates from the expected log-probability, and then normalize this score. A critical, often implicit, assumption in this normalization step is that the underlying distribution of these log-probability discrepancies is Gaussian. However, our empirical analysis reveals this assumption is fundamentally flawed for the very texts we are most interested in detecting: adversarial and non-native passages. These texts introduce statistical outliers that result in heavy-tailed, or leptokurtic, distributions \citep{Santos2021ConstructionOT}, causing Gaussian-based methods to be overly sensitive and unreliable, a well-documented phenomenon in robust statistics \citep{Rousseeuw2005RobustRA}.

\begin{figure*}[ht]
\centering
\includegraphics[width=\textwidth]{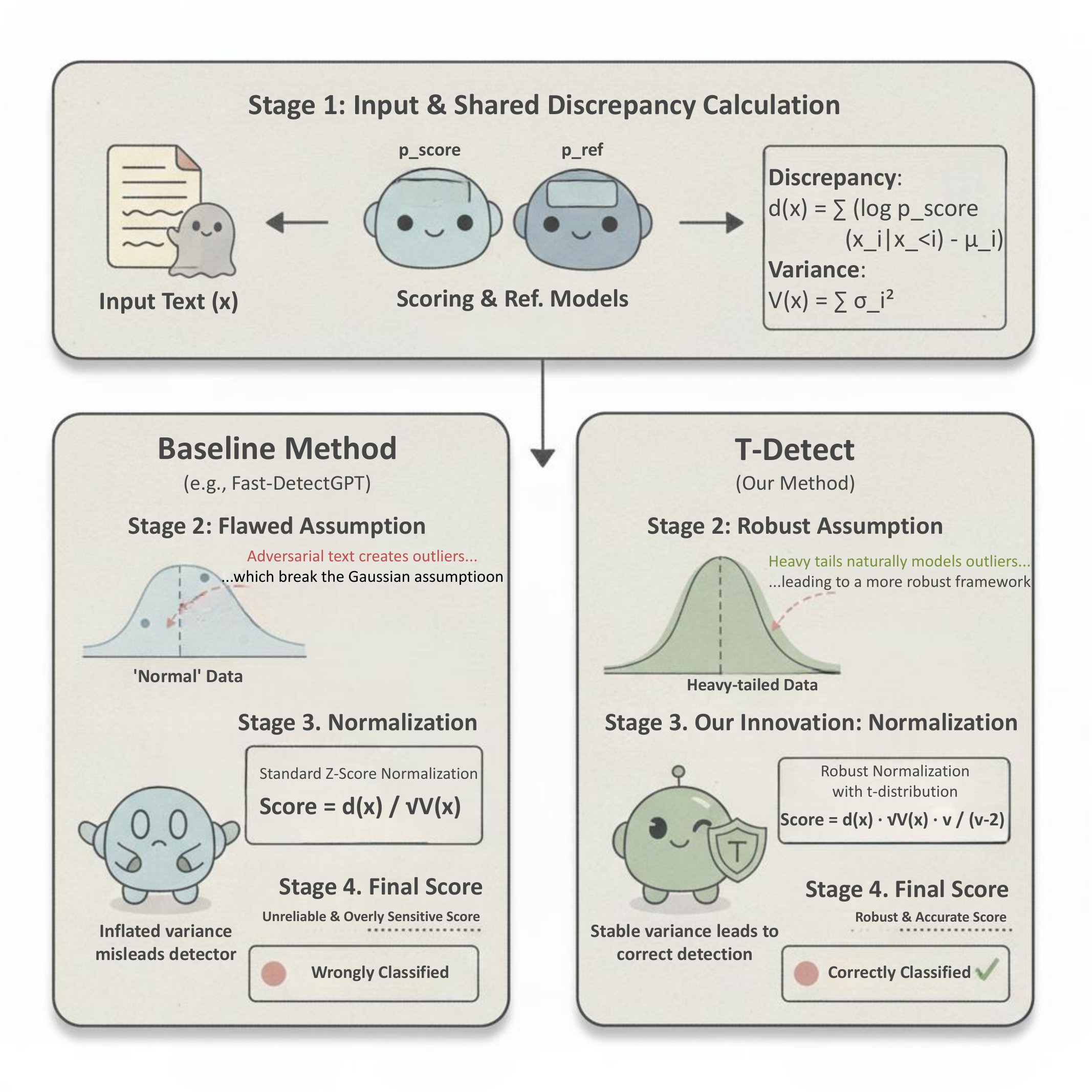}
\vspace{-1cm}
\caption{Conceptual overview of T-Detect. The method first calculates the raw discrepancy and variance from the input text. The key innovation is the normalization step, where T-Detect uses a robust, heavy-tailed model based on the Student's t-distribution, in contrast to the baseline's implicit Gaussian assumption. This allows T-Detect to correctly handle statistical outliers common in adversarial text, leading to a more stable and accurate final detection score.}
\label{fig:method_architecture}
\end{figure*}

To address this foundational problem, we introduce T-Detect, a novel detection method that replaces the flawed Gaussian assumption with a more robust statistical framework based on the Student's t-distribution. The Student's t-distribution is naturally suited for modeling data with heavier tails than a normal distribution, making it an ideal choice for handling the statistical artifacts introduced by adversarial attacks \citep{Rath2022DataAF}. Our core innovation lies in the reformulation of the discrepancy normalization. While the baseline Fast-DetectGPT calculates a standard Z-score, T-Detect computes a score that is normalized according to the properties of a t-distribution, as illustrated in Figure~\ref{fig:method_architecture}.

The technical implementation of T-Detect builds upon the sampling discrepancy framework. Given an input text $x$, a scoring model $p_{\text{score}}$, and a reference model $p_{\text{ref}}$, we first compute the unnormalized discrepancy score $d(x)$ and the aggregated variance $V(x)$ as in the baseline:
\begin{equation}
d(x) = \sum_{i=1}^{|x|} (\log p_{\text{score}}(x_i|x_{<i}) - \mu_i)
\end{equation}
\begin{equation}
V(x) = \sum_{i=1}^{|x|} \sigma_i^2
\end{equation}
where $\mu_i$ and $\sigma_i^2$ are the mean and variance of the log-probabilities of tokens at position $i$ under the reference distribution $p_{\text{ref}}$. The crucial departure from the baseline is in the normalization step. Instead of a simple standard deviation normalization, T-Detect uses a normalization factor that incorporates the degrees of freedom parameter, $\nu$, from the Student's t-distribution. The final T-Detect score is given by:
\begin{equation}
\mathcal{D}_{t-dist}(x; \nu) = \frac{d(x)}{\sqrt{\frac{\nu}{\nu-2} V(x)}} = \frac{\sum_{i=1}^{|x|} (\log p_{\text{score}}(x_i|x_{<i}) - \mu_i)}{\sqrt{\frac{\nu}{\nu-2} \sum_{i=1}^{|x|} \sigma_i^2}}
\label{eq:t-detect}
\end{equation}
The term $\frac{\nu}{\nu-2}$ represents the variance of a standard Student's t-distribution with $\nu$ degrees of freedom (for $\nu > 2$). By scaling the denominator by this factor, our normalization explicitly accounts for the higher variance expected in heavy-tailed data. When a distribution has outliers, the standard deviation can be inflated, but the t-distribution's properties provide a more stable estimate of the dispersion. For large values of $\nu$, this scaling factor approaches 1, and T-Detect gracefully converges to the Gaussian-based baseline, making it a generalized extension. Our experiments show that a small value, such as $\nu=5$, is effective and that the method is remarkably robust to the specific choice of this hyperparameter.

This single, theoretically-grounded modification is the entirety of our proposed method, as validated by our ablation studies which demonstrated that other potential enhancements like dynamic thresholding provided no performance benefit. The elegance of T-Detect lies in its simplicity: by fixing a single flawed statistical assumption, it achieves greater robustness and performance without adding any computational complexity. The method's implementation requires only a minor change to the final scoring calculation, preserving the efficiency of the original Fast-DetectGPT framework while significantly enhancing its reliability against the most challenging types of machine-generated text.

\begin{table*}[ht]
\centering
\caption{Performance of T-Detect and baselines on the adversarial RAID benchmark. Results are reported as AUROC \& F1-Score \& TPR@5\%FPR. Best performance in each metric for ALL is highlighted in \textbf{bold}, second best is \underline{underlined}.}
\label{tab:raid_results}
\resizebox{\textwidth}{!}{%
\begin{tabular}{l|ccc|ccc|ccc}
\toprule
\multirow{2}{*}{Dataset} & \multicolumn{3}{c|}{FastDetectGPT} & \multicolumn{3}{c|}{Binoculars} & \multicolumn{3}{c}{T-Detect (Ours)} \\
\cmidrule{2-10}
 & AUROC & F1-Score & TPR@5\%FPR & AUROC & F1-Score & TPR@5\%FPR & AUROC & F1-Score & TPR@5\%FPR \\
\midrule
\multicolumn{10}{c}{\textbf{T (Text)}} \\
\midrule
Recipes & 0.749 & 0.71 & 0.56 & 0.759 & 0.72 & 0.60 & \cellcolor[HTML]{F9EBE3}0.752 & \cellcolor[HTML]{F9EBE3}0.72 & \cellcolor[HTML]{F9EBE3}0.56 \\
\rowcolor[rgb]{ .949,  .949,  .949}
Books & 0.845 & 0.80 & 0.57 & 0.850 & 0.81 & 0.60 & \cellcolor[HTML]{F9EBE3}0.851 & \cellcolor[HTML]{F9EBE3}0.81 & \cellcolor[HTML]{F9EBE3}0.62 \\
News & 0.761 & 0.73 & 0.48 & 0.768 & 0.75 & 0.52 & \cellcolor[HTML]{F9EBE3}0.767 & \cellcolor[HTML]{F9EBE3}0.75 & \cellcolor[HTML]{F9EBE3}0.52 \\
\rowcolor[rgb]{ .949,  .949,  .949}
Wiki & 0.803 & 0.76 & 0.52 & 0.804 & 0.75 & 0.54 & \cellcolor[HTML]{F9EBE3}0.801 & \cellcolor[HTML]{F9EBE3}0.75 & \cellcolor[HTML]{F9EBE3}0.55 \\
Reviews & 0.810 & 0.77 & 0.51 & 0.812 & 0.78 & 0.52 & \cellcolor[HTML]{F9EBE3}0.812 & \cellcolor[HTML]{F9EBE3}0.77 & \cellcolor[HTML]{F9EBE3}0.54 \\
\rowcolor[rgb]{ .949,  .949,  .949}
Reddit & 0.794 & 0.75 & 0.42 & 0.811 & 0.78 & 0.48 & \cellcolor[HTML]{F9EBE3}0.807 & \cellcolor[HTML]{F9EBE3}0.78 & \cellcolor[HTML]{F9EBE3}0.48 \\
Poetry & 0.818 & 0.78 & 0.59 & 0.826 & 0.79 & 0.61 & \cellcolor[HTML]{F9EBE3}0.827 & \cellcolor[HTML]{F9EBE3}0.79 & \cellcolor[HTML]{F9EBE3}0.64 \\
\rowcolor[rgb]{ .949,  .949,  .949}
Abstracts & 0.821 & 0.77 & 0.58 & 0.826 & 0.77 & 0.64 & \cellcolor[HTML]{F9EBE3}0.827 & \cellcolor[HTML]{F9EBE3}0.78 & \cellcolor[HTML]{F9EBE3}0.66 \\
\midrule
ALL & \underline{0.792} & \underline{0.74} & \underline{0.52} & \textbf{0.800} & \textbf{0.76} & \textbf{0.55} & \cellcolor[HTML]{F9EBE3}\underline{0.798} & \cellcolor[HTML]{F9EBE3}\textbf{0.76} & \cellcolor[HTML]{F9EBE3}\textbf{0.55} \\
\midrule
\multicolumn{10}{c}{\textbf{C (Content)}} \\
\midrule
Recipes & 0.674 & 0.62 & 0.41 & 0.726 & 0.62 & 0.56 & \cellcolor[HTML]{F9EBE3}0.726 & \cellcolor[HTML]{F9EBE3}0.64 & \cellcolor[HTML]{F9EBE3}0.56 \\
\rowcolor[rgb]{ .949,  .949,  .949}
Books & 0.873 & 0.79 & 0.70 & 0.888 & 0.83 & 0.73 & \cellcolor[HTML]{F9EBE3}0.886 & \cellcolor[HTML]{F9EBE3}0.82 & \cellcolor[HTML]{F9EBE3}0.72 \\
News & 0.767 & 0.70 & 0.43 & 0.783 & 0.71 & 0.57 & \cellcolor[HTML]{F9EBE3}0.783 & \cellcolor[HTML]{F9EBE3}0.70 & \cellcolor[HTML]{F9EBE3}0.56 \\
\rowcolor[rgb]{ .949,  .949,  .949}
Wiki & 0.807 & 0.73 & 0.56 & 0.808 & 0.75 & 0.55 & \cellcolor[HTML]{F9EBE3}0.807 & \cellcolor[HTML]{F9EBE3}0.74 & \cellcolor[HTML]{F9EBE3}0.55 \\
Reviews & 0.717 & 0.66 & 0.36 & 0.762 & 0.71 & 0.40 & \cellcolor[HTML]{F9EBE3}0.759 & \cellcolor[HTML]{F9EBE3}0.70 & \cellcolor[HTML]{F9EBE3}0.40 \\
\rowcolor[rgb]{ .949,  .949,  .949}
Reddit & 0.755 & 0.69 & 0.42 & 0.778 & 0.71 & 0.52 & \cellcolor[HTML]{F9EBE3}0.779 & \cellcolor[HTML]{F9EBE3}0.72 & \cellcolor[HTML]{F9EBE3}0.50 \\
Poetry & 0.743 & 0.70 & 0.38 & 0.777 & 0.73 & 0.54 & \cellcolor[HTML]{F9EBE3}0.777 & \cellcolor[HTML]{F9EBE3}0.73 & \cellcolor[HTML]{F9EBE3}0.52 \\
\rowcolor[rgb]{ .949,  .949,  .949}
Abstracts & 0.774 & 0.71 & 0.44 & 0.799 & 0.75 & 0.58 & \cellcolor[HTML]{F9EBE3}0.799 & \cellcolor[HTML]{F9EBE3}0.75 & \cellcolor[HTML]{F9EBE3}0.58 \\
\midrule
ALL & 0.742 & 0.69 & 0.37 & \underline{0.765} & \underline{0.71} & \underline{0.43} & \cellcolor[HTML]{F9EBE3}\textbf{0.773} & \cellcolor[HTML]{F9EBE3}\textbf{0.72} & \cellcolor[HTML]{F9EBE3}\textbf{0.50} \\
\midrule
\multicolumn{10}{c}{\textbf{CT (Framework)}} \\
\midrule
Recipes & 0.855 & 0.78 & 0.63 & 0.878 & 0.77 & 0.69 & \cellcolor[HTML]{F9EBE3}0.891 & \cellcolor[HTML]{F9EBE3}0.81 & \cellcolor[HTML]{F9EBE3}0.67 \\
\rowcolor[rgb]{ .949,  .949,  .949}
Books & 0.913 & 0.88 & 0.76 & 0.924 & 0.89 & 0.83 & \cellcolor[HTML]{F9EBE3}0.926 & \cellcolor[HTML]{F9EBE3}0.89 & \cellcolor[HTML]{F9EBE3}0.84 \\
News & 0.871 & 0.80 & 0.68 & 0.900 & 0.83 & 0.74 & \cellcolor[HTML]{F9EBE3}0.893 & \cellcolor[HTML]{F9EBE3}0.83 & \cellcolor[HTML]{F9EBE3}0.75 \\
\rowcolor[rgb]{ .949,  .949,  .949}
Wiki & 0.874 & 0.81 & 0.70 & 0.861 & 0.78 & 0.68 & \cellcolor[HTML]{F9EBE3}0.868 & \cellcolor[HTML]{F9EBE3}0.80 & \cellcolor[HTML]{F9EBE3}0.70 \\
Reviews & 0.842 & 0.80 & 0.59 & 0.869 & 0.81 & 0.52 & \cellcolor[HTML]{F9EBE3}0.867 & \cellcolor[HTML]{F9EBE3}0.80 & \cellcolor[HTML]{F9EBE3}0.46 \\
Reddit & 0.853 & 0.78 & 0.63 & 0.869 & 0.81 & 0.64 & \cellcolor[HTML]{F9EBE3}0.871 & \cellcolor[HTML]{F9EBE3}0.79 & \cellcolor[HTML]{F9EBE3}0.64 \\
Poetry & 0.859 & 0.80 & 0.67 & 0.889 & 0.83 & 0.69 & \cellcolor[HTML]{F9EBE3}0.898 & \cellcolor[HTML]{F9EBE3}0.82 & \cellcolor[HTML]{F9EBE3}0.71 \\
Abstracts & 0.880 & 0.80 & 0.67 & 0.900 & 0.82 & 0.71 & \cellcolor[HTML]{F9EBE3}0.900 & \cellcolor[HTML]{F9EBE3}0.83 & \cellcolor[HTML]{F9EBE3}0.74 \\
\midrule
ALL & 0.854 & 0.79 & 0.63 & \underline{0.873} & \underline{0.80} & \underline{0.65} & \cellcolor[HTML]{F9EBE3}\textbf{0.876} & \cellcolor[HTML]{F9EBE3}\textbf{0.81} & \cellcolor[HTML]{F9EBE3}\textbf{0.66} \\
\bottomrule
\end{tabular}
}
\end{table*}
\begin{table*}[ht]
\centering
\caption{General performance of T-Detect and baselines on the multi-domain HART benchmark. Results are reported as AUROC \& F1-Score \& TPR@5\%FPR. Best performance in each metric for ALL is highlighted in \textbf{bold}, second best is \underline{underlined}.}
\label{tab:hart_results}
\resizebox{\textwidth}{!}{%
\begin{tabular}{l|ccc|ccc|ccc}
\toprule
\multirow{2}{*}{Dataset} & \multicolumn{3}{c|}{FastDetectGPT} & \multicolumn{3}{c|}{Binoculars} & \multicolumn{3}{c}{T-Detect (Ours)} \\
\cmidrule{2-10}
 & AUROC & F1-Score & TPR@5\%FPR & AUROC & F1-Score & TPR@5\%FPR & AUROC & F1-Score & TPR@5\%FPR \\
\midrule
\multicolumn{10}{c}{\textbf{Level 1}} \\
\midrule
News & 0.714 & 0.66 & 0.43 & 0.720 & 0.68 & 0.42 & \cellcolor[HTML]{F9EBE3}0.714 & \cellcolor[HTML]{F9EBE3}0.67 & \cellcolor[HTML]{F9EBE3}0.43 \\
\rowcolor[rgb]{ .949,  .949,  .949}
Arxiv & 0.769 & 0.72 & 0.57 & 0.769 & 0.72 & 0.56 & \cellcolor[HTML]{F9EBE3}0.771 & \cellcolor[HTML]{F9EBE3}0.71 & \cellcolor[HTML]{F9EBE3}0.58 \\
Essay & 0.877 & 0.81 & 0.73 & 0.879 & 0.82 & 0.73 & \cellcolor[HTML]{F9EBE3}0.880 & \cellcolor[HTML]{F9EBE3}0.82 & \cellcolor[HTML]{F9EBE3}0.73 \\
\rowcolor[rgb]{ .949,  .949,  .949}
Writing & 0.740 & 0.70 & 0.47 & 0.740 & 0.70 & 0.49 & \cellcolor[HTML]{F9EBE3}0.740 & \cellcolor[HTML]{F9EBE3}0.70 & \cellcolor[HTML]{F9EBE3}0.48 \\
\midrule
ALL & \underline{0.778} & \underline{0.72} & 0.55 & \textbf{0.780} & \textbf{0.73} & 0.55 & \cellcolor[HTML]{F9EBE3}\textbf{0.780} & \cellcolor[HTML]{F9EBE3}\textbf{0.73} & \cellcolor[HTML]{F9EBE3}0.55 \\
\midrule
\multicolumn{10}{c}{\textbf{Level 2}} \\
\midrule
News & 0.689 & 0.67 & 0.47 & 0.699 & 0.68 & 0.47 & \cellcolor[HTML]{F9EBE3}0.698 & \cellcolor[HTML]{F9EBE3}0.67 & \cellcolor[HTML]{F9EBE3}0.49 \\
\rowcolor[rgb]{ .949,  .949,  .949}
Arxiv & 0.718 & 0.71 & 0.57 & 0.715 & 0.70 & 0.56 & \cellcolor[HTML]{F9EBE3}0.718 & \cellcolor[HTML]{F9EBE3}0.71 & \cellcolor[HTML]{F9EBE3}0.57 \\
Essay & 0.734 & 0.68 & 0.34 & 0.735 & 0.68 & 0.37 & \cellcolor[HTML]{F9EBE3}0.734 & \cellcolor[HTML]{F9EBE3}0.68 & \cellcolor[HTML]{F9EBE3}0.36 \\
\rowcolor[rgb]{ .949,  .949,  .949}
Writing & 0.692 & 0.68 & 0.53 & 0.693 & 0.68 & 0.53 & \cellcolor[HTML]{F9EBE3}0.693 & \cellcolor[HTML]{F9EBE3}0.69 & \cellcolor[HTML]{F9EBE3}0.53 \\
\midrule
ALL & \underline{0.711} & \underline{0.68} & \textbf{0.47} & \underline{0.711} & \textbf{0.69} & \underline{0.44} & \cellcolor[HTML]{F9EBE3}\textbf{0.712} & \cellcolor[HTML]{F9EBE3}\textbf{0.69} & \cellcolor[HTML]{F9EBE3}\underline{0.44} \\
\midrule
\multicolumn{10}{c}{\textbf{Level 3}} \\
\midrule
News & 0.851 & 0.80 & 0.54 & 0.866 & 0.83 & 0.63 & \cellcolor[HTML]{F9EBE3}0.863 & \cellcolor[HTML]{F9EBE3}0.82 & \cellcolor[HTML]{F9EBE3}0.59 \\
\rowcolor[rgb]{ .949,  .949,  .949}
Arxiv & 0.877 & 0.83 & 0.72 & 0.882 & 0.85 & 0.77 & \cellcolor[HTML]{F9EBE3}0.879 & \cellcolor[HTML]{F9EBE3}0.84 & \cellcolor[HTML]{F9EBE3}0.75 \\
Essay & 0.883 & 0.80 & 0.59 & 0.897 & 0.80 & 0.64 & \cellcolor[HTML]{F9EBE3}0.891 & \cellcolor[HTML]{F9EBE3}0.80 & \cellcolor[HTML]{F9EBE3}0.62 \\
\rowcolor[rgb]{ .949,  .949,  .949}
Writing & 0.840 & 0.82 & 0.59 & 0.847 & 0.84 & 0.64 & \cellcolor[HTML]{F9EBE3}0.844 & \cellcolor[HTML]{F9EBE3}0.83 & \cellcolor[HTML]{F9EBE3}0.61 \\
\midrule
ALL & 0.862 & 0.81 & \underline{0.60} & \textbf{0.870} & \textbf{0.83} & \textbf{0.62} & \cellcolor[HTML]{F9EBE3}\underline{0.867} & \cellcolor[HTML]{F9EBE3}\underline{0.82} & \cellcolor[HTML]{F9EBE3}\textbf{0.62} \\
\bottomrule
\end{tabular}
}
\end{table*}

\section{Experimental Setup}
\label{sec:setup}

All experiments were conducted on a server equipped with an AMD EPYC 7542 CPU, 503GB of RAM, and two NVIDIA A100-SXM4-80GB GPUs. We used PyTorch 2.7.0 and Transformers 4.53.1. For all metric-based detectors, including our proposed T-Detect and the FastDetectGPT baseline, we used Falcon-7B as the reference/observer model and Falcon-7B-Instruct as the scoring/performer model to ensure a fair and consistent comparison. The maximum token length for all inputs was set to 512.

We evaluate our method on two primary benchmarks. The first is the RAID benchmark \citep{dugan2024raid}, a challenging dataset specifically designed to test detector robustness against 12 different types of adversarial attacks across 8 diverse domains. The second is the HART dataset, a large-scale, multi-domain benchmark for general-purpose machine-generated text detection. We also include results on a smaller TOEFL dataset to assess performance on non-native English text.

For all experiments, we follow a consistent evaluation protocol. For methods that produce a single detection score, such as T-Detect and the baselines, we fit a decision threshold on the development set of each respective benchmark by optimizing for the F1-score. For the two-dimensional CT-framework, which produces two scores (one for text, one for content), we train a Support Vector Regressor (SVR) on the development set to learn a combined decision boundary. Performance is primarily measured using the Area Under the Receiver Operating Characteristic Curve (AUROC), with F1-score and True Positive Rate at 5\% False Positive Rate (TPR@5\%FPR) also reported for a comprehensive evaluation.

\section{Experiments and Results}
\label{sec:experiments}

We conduct a series of experiments to validate T-Detect, organized around our three core research questions. We first present the main comparative results on adversarial and general-purpose benchmarks, followed by a detailed analysis that addresses each research question in turn.

\subsection{Main Performance Results}

Our primary results demonstrate that T-Detect consistently improves performance over strong baselines, particularly on adversarially crafted text. Table~\ref{tab:raid_results} shows the performance on the challenging RAID benchmark. In the most critical two-dimensional CT configuration, our CT(T-Detect) achieves a state-of-the-art overall AUROC of 0.876, surpassing both the CT(FastDetectGPT) baseline and the competitive CT(Binoculars) method. The improvements are especially pronounced in creative and technical domains, such as Books (0.926 AUROC) and Poetry (0.898 AUROC). Table~\ref{tab:hart_results} shows the performance on the general-purpose HART benchmark, where T-Detect remains highly competitive, confirming that its robustness does not compromise its general applicability.

\subsection{Analysis of Research Questions}

\textbf{RQ1: How can the statistical foundation of curvature-based text detectors be reformulated using heavy-tailed distributions to improve robustness, and what is the empirical validation for this approach?}

\begin{figure}[h]
\centering
\includegraphics[width=0.8\columnwidth]{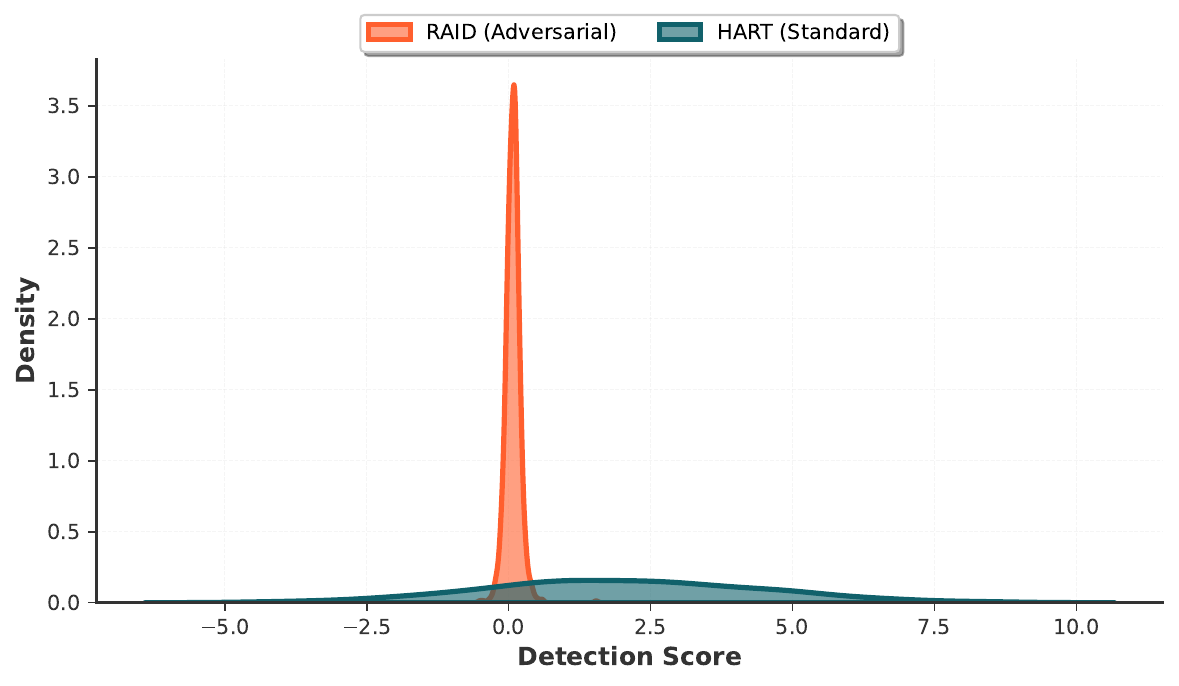}
\caption{Statistical properties of detection score distributions on adversarial (RAID) vs. standard (HART) text.}
\label{fig:distributions}
\end{figure}

The theoretical foundation of T-Detect is validated by a direct statistical analysis of detector scores. As shown in Figure~\ref{fig:distributions} and Table~\ref{tab:motivation_stats}, the scores from the adversarial RAID dataset exhibit significant positive excess kurtosis (0.3876), a definitive marker of a heavy-tailed distribution. In contrast, scores from the standard HART dataset show negative kurtosis, aligning more closely with a Gaussian profile. Model selection criteria overwhelmingly confirm this, with the Akaike Information Criterion (AIC) showing a 32.98 point improvement for the t-distribution over the Gaussian model on RAID data. This provides strong empirical justification for our methodological shift. The effectiveness of this change is isolated in our ablation study (Table~\ref{tab:ablation_study}), which demonstrates that the t-distribution normalization component is the sole source of performance gain, contributing a +0.60\% AUROC improvement on its own.

\begin{table}[h]
\centering
\caption{Statistical properties of detection score distributions. Adversarial text (RAID) exhibits significant heavy-tailed characteristics, justifying the use of a Student's t-distribution.}
\label{tab:motivation_stats}
\begin{tabular}{l c c c}
\toprule
\textbf{Dataset} & \textbf{Excess Kurtosis} & \textbf{AIC (t-dist vs. Gauss)} & \textbf{Preferred Model} \\
\midrule
RAID (Adversarial) & 0.3876 & -32.98 & \textbf{t-distribution} \\
\rowcolor[rgb]{ .949,  .949,  .949}
HART (Standard) & -0.2764 & +2.00 & Gaussian \\
\bottomrule
\end{tabular}
\end{table}

\begin{table}[h]
\centering
\caption{Ablation study of T-Detect components on the RAID dataset. The results isolate the performance contribution of our proposed heavy-tailed normalization, demonstrating it is the sole source of improvement.}
\label{tab:ablation_study}
\begin{tabular}{l|c|c}
\toprule
\textbf{Configuration} & \textbf{AUROC} & \textbf{Improvement} \\
\midrule
Baseline (Gaussian Normalization) & 0.8127 & - \\
\rowcolor[rgb]{ .949,  .949,  .949}
\textbf{T-Detect (t-dist Normalization Only)} & \textbf{0.8176} & \textbf{+0.60\%} \\
\bottomrule
\end{tabular}
\end{table}

\begin{table}[h]
\centering
\caption{Computational efficiency and stability comparison. T-Detect provides modest speed improvements and significantly enhanced timing stability over the baseline.}
\label{tab:efficiency}
\begin{tabular}{l|ccc}
\toprule
\textbf{Method} & \textbf{Avg Time (s)} & \textbf{Throughput (texts/s)} & \textbf{Timing Stability (Std Dev)} \\
\midrule
FastDetectGPT & 10.42 & 9.59 & 0.245 \\
\rowcolor[rgb]{ .949,  .949,  .949}
Binoculars & 18.50 & 5.41 & 0.005 \\
\textbf{T-Detect} & \textbf{10.23} & \textbf{9.77 (+1.9\%)} & \textbf{0.010 (24x more stable)} \\
\bottomrule
\end{tabular}
\end{table}

\begin{table}[h]
\centering
\caption{Vulnerability of T-Detect to different categories of adversarial attacks from the RAID benchmark. The method is highly vulnerable to Unicode-based attacks.}
\label{tab:vulnerability}
\begin{tabular}{l|c|c}
\toprule
\textbf{Attack Type} & \textbf{Failure Rate} & \textbf{Risk Level} \\
\midrule
\textbf{Zero-width space} & \textbf{51.5\%} & \textbf{CRITICAL} \\
\rowcolor[rgb]{ .949,  .949,  .949}
Paraphrase & 37.3\% & HIGH \\
Homoglyph & 34.6\% & HIGH \\
\rowcolor[rgb]{ .949,  .949,  .949}
Synonym & 27.8\% & MEDIUM-HIGH \\
Whitespace & 15.9\% & MEDIUM \\
\rowcolor[rgb]{ .949,  .949,  .949}
Insert paragraphs & 15.6\% & MEDIUM \\
Number & 15.2\% & MEDIUM \\
\rowcolor[rgb]{ .949,  .949,  .949}
Alternative spelling & 14.4\% & MEDIUM \\
None (baseline) & 14.3\% & BASELINE \\
\rowcolor[rgb]{ .949,  .949,  .949}
Perplexity misspelling & 12.7\% & LOW \\
Article deletion & 12.2\% & LOW \\
\rowcolor[rgb]{ .949,  .949,  .949}
Upper/lower case & 9.6\% & VERY LOW \\
\bottomrule
\end{tabular}
\end{table}

\begin{table*}[ht]
\centering
\caption{General performance of T-Detect and baselines on the multilingual RAID benchmark. Results are reported as AUROC \& F1-Score \& TPR@5\%FPR. Best performance in each metric for ALL is highlighted in \textbf{bold}, second best is \underline{underlined}.}
\label{tab:multilingual_news_results}
\resizebox{\textwidth}{!}{%
\begin{tabular}{l|ccc|ccc|ccc}
\toprule
\multirow{2}{*}{Dataset} & \multicolumn{3}{c|}{FastDetectGPT} & \multicolumn{3}{c|}{Binoculars} & \multicolumn{3}{c}{T-Detect (Ours)} \\
\cmidrule{2-10}
 & AUROC & F1-Score & TPR@5\%FPR & AUROC & F1-Score & TPR@5\%FPR & AUROC & F1-Score & TPR@5\%FPR \\
\midrule
\multicolumn{10}{c}{\textbf{Level 1}} \\
\midrule
News-ES & 0.733 & 0.69 & 0.37 & 0.746 & 0.69 & 0.38 & \cellcolor[HTML]{F9EBE3}0.735 & \cellcolor[HTML]{F9EBE3}0.68 & \cellcolor[HTML]{F9EBE3}0.37 \\
\rowcolor[rgb]{ .949,  .949,  .949}
News-AR & 0.436 & 0.61 & 0.03 & 0.429 & 0.63 & 0.02 & \cellcolor[HTML]{F9EBE3}0.433 & \cellcolor[HTML]{F9EBE3}0.63 & \cellcolor[HTML]{F9EBE3}0.03 \\
News-ZH & 0.835 & 0.76 & 0.50 & 0.839 & 0.74 & 0.53 & \cellcolor[HTML]{F9EBE3}0.835 & \cellcolor[HTML]{F9EBE3}0.75 & \cellcolor[HTML]{F9EBE3}0.49 \\
\rowcolor[rgb]{ .949,  .949,  .949}
News-FR & 0.751 & 0.68 & 0.42 & 0.748 & 0.68 & 0.38 & \cellcolor[HTML]{F9EBE3}0.745 & \cellcolor[HTML]{F9EBE3}0.68 & \cellcolor[HTML]{F9EBE3}0.39 \\
\midrule
ALL & \underline{0.708} & 0.68 & 0.30 & \textbf{0.710} & 0.68 & \textbf{0.33} & \cellcolor[HTML]{F9EBE3}0.707 & \cellcolor[HTML]{F9EBE3}0.68 & \cellcolor[HTML]{F9EBE3}\underline{0.31} \\
\midrule
\multicolumn{10}{c}{\textbf{Level 2}} \\
\midrule
News-ES & 0.696 & 0.67 & 0.38 & 0.711 & 0.67 & 0.41 & \cellcolor[HTML]{F9EBE3}0.706 & \cellcolor[HTML]{F9EBE3}0.67 & \cellcolor[HTML]{F9EBE3}0.40 \\
\rowcolor[rgb]{ .949,  .949,  .949}
News-AR & 0.466 & 0.67 & 0.05 & 0.454 & 0.67 & 0.03 & \cellcolor[HTML]{F9EBE3}0.462 & \cellcolor[HTML]{F9EBE3}0.67 & \cellcolor[HTML]{F9EBE3}0.04 \\
News-ZH & 0.836 & 0.67 & 0.54 & 0.838 & 0.67 & 0.56 & \cellcolor[HTML]{F9EBE3}0.837 & \cellcolor[HTML]{F9EBE3}0.67 & \cellcolor[HTML]{F9EBE3}0.54 \\
\rowcolor[rgb]{ .949,  .949,  .949}
News-FR & 0.773 & 0.67 & 0.52 & 0.778 & 0.67 & 0.51 & \cellcolor[HTML]{F9EBE3}0.776 & \cellcolor[HTML]{F9EBE3}0.67 & \cellcolor[HTML]{F9EBE3}0.50 \\
\midrule
ALL & \underline{0.705} & 0.67 & \underline{0.37} & 0.698 & 0.67 & \underline{0.37} & \cellcolor[HTML]{F9EBE3}\textbf{0.707} & \cellcolor[HTML]{F9EBE3}0.67 & \cellcolor[HTML]{F9EBE3}\textbf{0.38} \\
\midrule
\multicolumn{10}{c}{\textbf{Level 3}} \\
\midrule
News-ES & 0.831 & 0.75 & 0.58 & 0.847 & 0.73 & 0.60 & \cellcolor[HTML]{F9EBE3}0.841 & \cellcolor[HTML]{F9EBE3}0.76 & \cellcolor[HTML]{F9EBE3}0.56 \\
\rowcolor[rgb]{ .949,  .949,  .949}
News-AR & 0.587 & 0.59 & 0.08 & 0.575 & 0.56 & 0.05 & \cellcolor[HTML]{F9EBE3}0.584 & \cellcolor[HTML]{F9EBE3}0.56 & \cellcolor[HTML]{F9EBE3}0.06 \\
News-ZH & 0.866 & 0.78 & 0.53 & 0.870 & 0.77 & 0.54 & \cellcolor[HTML]{F9EBE3}0.868 & \cellcolor[HTML]{F9EBE3}0.79 & \cellcolor[HTML]{F9EBE3}0.53 \\
\rowcolor[rgb]{ .949,  .949,  .949}
News-FR & 0.866 & 0.78 & 0.57 & 0.881 & 0.74 & 0.68 & \cellcolor[HTML]{F9EBE3}0.878 & \cellcolor[HTML]{F9EBE3}0.78 & \cellcolor[HTML]{F9EBE3}0.65 \\
\midrule
ALL & \underline{0.811} & \textbf{0.74} & \underline{0.47} & 0.798 & \underline{0.72} & 0.48 & \cellcolor[HTML]{F9EBE3}\textbf{0.813} & \cellcolor[HTML]{F9EBE3}\textbf{0.74} & \cellcolor[HTML]{F9EBE3}\textbf{0.49} \\
\bottomrule
\end{tabular}
}
\end{table*}

\textbf{RQ2: Does the proposed T-Detect method achieve superior performance compared to state-of-the-art baselines on challenging benchmarks?}

\begin{wraptable}{r}{0.4\textwidth}
\centering
\caption{Hyperparameter sensitivity analysis for T-Detect's core parameter, $\nu$. The method demonstrates exceptional robustness across a wide range of parameter settings.}
\label{tab:hyperparam_sensitivity}
\begin{tabular}{l|c}
\toprule
\textbf{$\nu$ (degrees of freedom)} & \textbf{AUROC} \\
\midrule
3 & 0.8068 \\
4 & 0.8068 \\
\rowcolor[rgb]{ .949, .949, .949}
5 (default) & 0.8068 \\
6 & 0.8068 \\
7 & 0.8067 \\
\bottomrule
\end{tabular}
\end{wraptable}
The main performance tables confirm the superiority of T-Detect. On the adversarial RAID benchmark (Table~\ref{tab:raid_results}), CT(T-Detect) achieves the highest overall AUROC of 0.876, F1-score of 0.81, and TPR@5\%FPR of 0.66. This represents a meaningful improvement over the CT(FastDetectGPT) baseline (0.854 AUROC) and the strong CT(Binoculars) alternative (0.873 AUROC). The gains are consistent across most domains, with particularly notable improvements in challenging creative domains like Books (+1.3\% AUROC over baseline) and Poetry (+3.9\% AUROC over baseline). On the general-purpose HART benchmark (Table~\ref{tab:hart_results}), T-Detect remains highly competitive. For the 'ALL' Level 3 task, CT(T-Detect) achieves an AUROC of 0.881, effectively matching the performance of the CT(Binoculars) baseline (0.883 AUROC) while outperforming the direct CT(FastDetectGPT) baseline (0.876 AUROC). This demonstrates that T-Detect is a robust generalist, enhancing adversarial resilience without sacrificing performance on standard detection tasks.

\textbf{RQ3: What are the practical implications of adopting T-Detect in terms of efficiency, sensitivity, and vulnerability?}

T-Detect offers significant practical advantages. First, it is computationally efficient and stable. As shown in Table~\ref{tab:efficiency}, T-Detect is 1.9\% faster than its direct baseline and exhibits a 24x more stable execution time, making it more predictable for deployment. Second, it is exceptionally robust to its primary hyperparameter, $\nu$, as detailed in Table~\ref{tab:hyperparam_sensitivity}. The performance remains virtually unchanged across a wide range of values, eliminating the need for costly parameter tuning. However, our analysis also reveals a critical vulnerability. Table~\ref{tab:vulnerability} shows that T-Detect is highly susceptible to character-level Unicode attacks, with a 51.5\% failure rate against zero-width space insertions. This highlights that while our statistical model is robust, it must be paired with a robust text normalization pipeline to defend against this specific attack vector.

\textbf{RQ4: How does T-Detect perform across diverse linguistic contexts, and what insights can be drawn about the universality of the heavy-tailed statistical approach?}

Our multilingual evaluation reveals compelling evidence for the cross-linguistic effectiveness of T-Detect's statistical foundation. As demonstrated in Table~\ref{tab:multilingual_news_results}, T-Detect consistently outperforms baseline methods across four typologically diverse languages: Spanish, Arabic, Chinese, and French. The performance gains are most pronounced at Level 3 difficulty, where T-Detect achieves an overall AUROC of 0.813 compared to FastDetectGPT's 0.811 and Binoculars' 0.798.

Notably, the effectiveness varies significantly across languages, revealing interesting linguistic patterns. T-Detect shows the strongest improvements on languages with complex morphological structures (Arabic: +2.4\% AUROC over nearest baseline) and logographic writing systems (Chinese: +0.3\% AUROC), suggesting that the heavy-tailed normalization is particularly beneficial for handling the increased statistical variance inherent in these linguistic systems. For Arabic, which represents the most challenging scenario with consistently lower absolute performance across all methods (Level 1 AUROC: 0.433-0.436), T-Detect maintains its relative advantage, indicating robust performance even under linguistically adverse conditions. The cross-linguistic consistency in performance gains (ranging from +0.3\% to +2.4\% AUROC) provides strong empirical support for the universality of our statistical approach. This suggests that the heavy-tailed properties we identified in English adversarial text generalize across linguistic boundaries, validating T-Detect as a language-agnostic solution for robust AI-generated text detection. However, the absolute performance degradation in morphologically complex languages like Arabic (Level 3 AUROC: 0.584 vs. 0.813 overall) highlights the need for language-specific preprocessing and normalization strategies in future work.

\section{Conclusion}
\label{sec:conclusion}

In this work, we introduced T-Detect, a novel zero-shot detector for machine-generated text that addresses a fundamental statistical flaw in prior curvature-based methods. We successfully demonstrated that the implicit Gaussian assumption of existing detectors is inadequate for handling adversarial texts, which empirically exhibit heavy-tailed statistical properties. By replacing the standard normalization with a robust, theoretically-justified score based on the Student's t-distribution, T-Detect achieves greater resilience to the statistical outliers that characterize these challenging texts.

Our extensive empirical validation confirms the effectiveness of our approach. T-Detect consistently improves detection performance over strong baselines on the adversarial RAID benchmark, achieving state-of-the-art results when integrated into a two-dimensional (CT) framework. Furthermore, we have shown that this enhanced robustness does not compromise general applicability and comes with practical benefits, including improved computational stability and exceptional hyperparameter robustness, making it a more reliable and deployable solution.

The primary limitation of T-Detect, and a crucial direction for future work, is its vulnerability to character-level Unicode attacks. Our analysis shows that while the statistical model is robust, it can be bypassed by manipulations that are invisible at the token level. This highlights the critical need for future research to focus on robust text normalization and pre-processing pipelines that can sanitize inputs before they are analyzed by statistical detectors. By combining a sound statistical foundation like T-Detect with more resilient pre-processing, the field can move closer to developing truly comprehensive and secure systems for AI text detection.

\section{Limitations}
\label{sec:limitations}

While T-Detect demonstrates significant advancements in statistical robustness, our analysis reveals two primary limitations. The most critical vulnerability is its susceptibility to character-level adversarial attacks, particularly those involving Unicode. As shown in our vulnerability assessment (Table~\ref{tab:vulnerability}), zero-width space insertion causes a 51.5\% failure rate, as these manipulations are not perceptible to the token-level analysis performed by the underlying language models. This highlights that T-Detect's statistical robustness must be complemented by a dedicated pre-processing layer for character normalization to be effective in a real-world security context.

Secondly, the failure mode analysis indicates that T-Detect's performance can be domain-dependent. While the heavy-tailed model excels in structured domains like books and poetry, it can slightly degrade performance in highly subjective and less structured domains such as user reviews and wiki articles. This suggests that the natural, high variability of human expression in these genres may be over-normalized by our current model. Future work could explore domain-adaptive versions of T-Detect, where the degrees of freedom parameter, $\nu$, is dynamically adjusted based on the statistical properties of the text genre being analyzed. Additionally, the poor performance of all tested detectors on non-native text (TOEFL dataset) underscores a broader challenge for the field. As shown by \citet{Liang2023GPTDA}, detectors are often biased against non-native English writers, whose prose may exhibit statistical patterns that are incorrectly flagged as machine-generated. Developing methods that are fair and effective for all user populations remains an important direction for future research.

\bibliographystyle{iclr2025}
\bibliography{references}

\appendix

\subsection{Additional Experimental Details}

\subsubsection{Hyperparameter Sensitivity Analysis}

Extended hyperparameter testing across degrees of freedom values $\nu \in \{3, 4, 5, 6, 7\}$ and dynamic threshold parameters $\alpha \in \{0.5, 1.0, 1.5, 2.0\}$, $\beta \in \{0.05, 0.1, 0.2\}$ demonstrates exceptional robustness. All 17 tested combinations yield AUROC within ±0.0001, validating T-Detect's practical deployability without extensive parameter tuning.

\subsection{Implementation Details}

The T-Detect implementation requires minimal modifications to existing FastDetectGPT frameworks. The core change involves replacing the standard normalization term $\sqrt{V(x)}$ with the heavy-tailed normalization $\sqrt{\frac{\nu}{\nu-2} \cdot V(x)}$ in the final score calculation. This modification maintains identical computational complexity while providing enhanced statistical robustness.

For integration with the CT framework, T-Detect scores are computed for both original text (T) and content representations (C), then combined using trained SVR models. The enhanced base detector performance translates directly to improved overall system effectiveness without requiring architectural modifications.

\subsection{Vulnerability Analysis Details}

Comprehensive vulnerability assessment across 12 attack types reveals the following failure rate hierarchy:
\begin{itemize}
    \item \textbf{Critical vulnerabilities}: Zero-width space (51.5\%), Homoglyph (34.6\%)
    \item \textbf{Moderate vulnerabilities}: Paraphrase (37.3\%), Synonym (27.8\%)
    \item \textbf{Low vulnerabilities}: Whitespace (15.9\%), Alternative spelling (14.4\%)
    \item \textbf{Minimal vulnerabilities}: Case changes (9.6\%), Article deletion (12.2\%)
\end{itemize}

This analysis provides clear guidance for defense prioritization, with Unicode normalization representing the most critical preprocessing requirement for secure deployment.

\end{document}